\documentclass[11pt,a4paper]{article}
\usepackage[hyperref]{conll2018}
\usepackage{times}
\usepackage{latexsym}

\usepackage{url}

\usepackage{multirow}

\usepackage{booktabs}

\usepackage[english]{babel}
\usepackage[utf8x]{inputenc}
\usepackage[T1]{fontenc}
\usepackage{microtype}

\usepackage{amsmath}
\usepackage{graphicx}
\usepackage{amssymb}
\usepackage{pgfplots, tikz}
\usepgfplotslibrary{groupplots}
\usepackage[colorinlistoftodos]{todonotes}
\usepackage{algpseudocode}
\usepackage{subcaption}
\pgfplotsset{compat=newest}
\usetikzlibrary{decorations.pathmorphing}

\aclfinalcopy 

\title{Lessons learned in multilingual grounded language learning}

\author{Ákos Kádár \\
	Tilburg University \\
   {\tt a.kadar@uvt.nl}\\\And
   Desmond Elliott\footnotemark[1] \\
	University of Copenhagen \\
   {\tt de@di.ku.dk}\\\And
   Marc-Alexandre Côté \\
   Microsoft Research Montreal\\
      {\tt macote@microsoft.com} \\\AND
      Grzegorz Chrupała\\
      Tilburg University \\
      {\tt g.chrupala@uvt.nl } \\\And
      Afra Alishahi\\
      Tilburg University \\
      {\tt a.alishahi@uvt.nl}\\
}

\date{}

\begin{document}
\maketitle

\begin{abstract}

Recent work has shown how to learn better visual-semantic embeddings by leveraging image descriptions in more than one language. Here, we investigate in detail which conditions affect the performance of this type of grounded language learning model. We show that multilingual training improves over bilingual training, and that low-resource languages benefit from training with higher-resource languages. We demonstrate that a multilingual model can be trained equally well on either translations or comparable sentence pairs, and that annotating the same set of images in multiple language enables further improvements via an additional caption-caption ranking objective.

\end{abstract}

\section{Introduction}

\renewcommand{\thefootnote}{\fnsymbol{footnote}}
\footnotetext[1]{Work carried out at the University of Edinburgh.}
\renewcommand{\thefootnote}{\arabic{footnote}}

Multimodal representation learning is largely motivated by evidence of perceptual grounding in human concept acquisition and representation \cite{barsalou2003grounding}. 
It has been shown that visually grounded word and sentence-representations  \cite{kiela2014improving,baroni2016grounding,elliott2017imagination,kiela2017learning,yoo2017improving} 
improve performance on the downstream tasks of paraphrase identification, semantic entailment, and multimodal machine translation
\cite{dolan2004unsupervised,marelli2014sick,specia-EtAl:2016:WMT}. 
Multilingual sentence representations have also been successfully applied to many-languages-to-one character-level machine translation \cite{chung2016character} and multilingual dependency parsing \cite{ammar2016many}. 

%
\begin{figure*}
  \begin{center}    
    \includegraphics[width=0.4\textwidth]{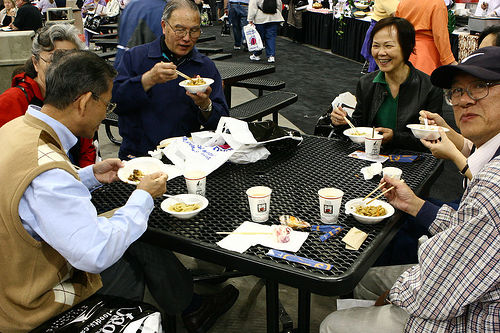}
  \end{center}
  \vspace{1em}  
  \begin{subfigure}[t]{0.5\textwidth}
    \textbf{En:} A group of people are eating noodles.\\[0.5ex]
    \textbf{De:} Eine Gruppe von Leuten isst Nudeln.\\[0.5ex]
    \textbf{Fr:} Un groupe de gens mangent des nouilles.\\[0.5ex]
    \textbf{Cs:} Skupina lidí jedí nudle.\\
    \vspace{2em}
    \caption{A translation tuple}
  \end{subfigure}
  \begin{subfigure}[t]{0.5\textwidth}
    \textbf{En:} Several asian people eating around a table.\\[0.5ex]
    \textbf{De:} Drei Männer und zwei Frauen südostasiatischen Aussehens sitzen, aus Schälchen essend, an einem schwarzen, Tisch, auf dem sich u.a. auch Pappbecher und eine Tasche befinden, im Hintergrund sind weitere Personen und Tische.\footnotemark\\
    \caption{A comparable pair}
  \end{subfigure}
  \caption{An example taken from the {\it Translation} and {\it Comparable} portions of the Multi30K dataset. The translation portion (a) contains professional translations of the English captions into German, French, and Czech. The comparable portion (b) consists of five independently crowdsourced English and German descriptions, given only the image. Note that the sentences in (b) convey different information from the English--German translation pair in (a).}\label{fig:data:example}
\end{figure*}

Recently, \citet{gella2017image} proposed to learn both bilingual and multimodal sentence representations using images paired with captions independently collected in English and German. Their results show that bilingual training improves image-sentence ranking performance over a monolingual baseline, and it improves performance on semantic textual similarity benchmarks \citep{agirre2014semeval,agirre2015semeval}. These findings suggest that it may be beneficial to consider another language as another {\it modality} in a monolingual grounded language learning model. In the grounded learning scenario, descriptions of an image in multiple languages can be considered as multiple views of the same or closely related data. These additional views can help overcome the problems of data sparsity, and have practical implications for efficiently collecting image-text datasets in different languages. In real-life applications, many tasks and domains can involve code switching \cite{barman2014code}, which is easier to deal with using a multilingual model. Furthermore, it is more convenient to maintain a single multilingual system than one system for each considered language. However, there is a need for a systematic exploration of the 
conditions under which it is useful to add additional views of the data. We investigate the impact of the following conditions on the performance of a multilingual grounded language learning model in sentence and image retrieval tasks:
\begin{description}
\item[Additional languages.] 
Multilingual models have not been explored yet in a multimodal setting. We investigate the contribution of adding more than one language by performing bilingual experiments on English and German (Section~\ref{sec:bi}) as well as adding French and Czech captioned images (Section~\ref{sec:multi}).

\item[Data alignment:] We assess the performance of a multilingual models trained using either captions that are translations of each other, or captions that are independently collected in different languages for the same set of images. The two scenarios are illustrated in Figure~\ref{fig:data:example}. Additionally we consider the setup when non-overlapping sets of images and their captions are collected in different languages. Such disjoint settings have been explored in pivot-based multimodal representation learning \cite{funaki2015image,rajendran2015bridge} or zero-shot multi-modal machine translation \cite{nakayama2017zero}. 
We compare translated vs.\ independently collected captions in Sections~\ref{sec:bitrans} and \ref{sec:multitrans}, and overlapping vs.\ disjoint images in Section~\ref{sec:bioverlap}. 

\item[High-to-low resource transfer:] In Section~\ref{sec:multitransfer} we investigate whether low-resource languages benefit from jointly training on larger data sets from higher-resource languages. This type of transfer has previously been shown to be effective in machine translation \cite[e.g.,][]{zoph2016transfer}.

\item[Training objective:] In addition to learning to map images to sentences, we study the effect of also learning relationships between captions of the same image in different languages \citet{gella2017image}. We assess the contribution of such a caption--caption ranking objective throughout our experiments. 
\end{description}
Our results show that multilingual joint training improves upon bilingual joint training, and that grounded sentence representations for a low-resource language can be substantially improved with data from different high-resource languages. Our results suggest that independently-collected captions are more useful than translated captions, for the task of learning multilingual multimodal sentence embeddings. Finally, we recommend to collect captions for the same set of images in multiple languages, due to the benefits of the additional caption--caption ranking objective function.

\section{Related work}

\footnotetext{Gloss: Three men and two women with a South-East Asian appearance eat out of bowls at a black table, on which there are, among other things, paper cups and a bag; in the background there are other people and tables.}

Learning visually grounded word-representations has been an active area of research in the fields of multi-modal semantics and cross-situational word-learning.
Such perceptually-grounded word representations have been shown to lead to higher correlation with human judgements on word-similarity benchmarks such as WordSim353 \cite{finkelstein2001placing} or SimLex999 \cite{hill2015simlex} compared to uni-modal representations \cite{kadar2015learning,bruni2014multimodal,kiela2014learning}. 

Grounded representations of sentences that are learned from image--caption data sets also improve performance on a number of sentence-level tasks \cite{kiela2017learning,yoo2017improving} when used as additional features to skip-thought vectors \cite{kiros2015skip}. 
The model architectures used for these studies  have the same overall structure as our model and coincide with image--sentence retrieval systems \cite{kiros2014unifying,karpathy2015deep}: a pre-trained CNN is fixed or fine-tuned as image feature extractor, followed by a learned transformation, while sentence representations are learned by a randomly initialized recurrent neural network. 
These models are trained to push the true image--caption pairs closer together, and the false image--caption pairs further from each other, in a joint embedding space. 

In addition to learning grounded representations for image-sentence ranking, joint vision and language systems have been proposed to solve a wide range of tasks across modalities such as image captioning \cite{mao2014deep,vinyals2015show,xu2015show}, visual question answering \cite{antol2015vqa,fukui2016multimodal,jabri2016revisiting}, 
text-to-image synthesis \cite{reed2016generative} 
and multimodal machine translation \cite{biblio732461325667051835,elliott2017imagination}.

Our work is also closely related to multilingual joint representation learning. 
In this scenario, a single model is trained to solve a task across multiple languages. 
\citet{ammar2016many} train a multilingual dependency parser on the Universal Dependencies treebank \cite{nivre2015universal} and show that on average the single multilingual model outperforms the monolingual baselines.  
\citet{johnson2016google} present a zero-shot neural machine translation model that is 
jointly trained on language pairs
$A \leftrightarrow B$ and $B \leftrightarrow C$ and show that the model is capable of performing well on the unseen
language pair $A \leftrightarrow C$. 
\citet{lee2017fully} find that jointly training a many-languages-to-one translation model on unsegmented character sequences improves BLEU scores compared to monolingual training. 
They also show evidence that the model can handle intra-sentence code-switching. 
\citet{peters2017massively} train a multilingual sequence-to-sequence translation architecture on grapheme-to-phoneme conversion using more than 300 languages. 
They report better performance when adding multiple languages, even those which are not present in the test data. 
Finally, massively multilingual language representations trained on over 900 languages have been shown to resemble language families \cite{ostling2016continuous} and can successfully predict linguistic typology features \cite{malaviya2017learning}. 

In the vision and language domain, multilingual-multimodal sentence representation learning has been limited so far to two languages. The joint training of models on English and German data has been shown to outperform monolingual baselines on image-sentence ranking and semantic textual similarity tasks \cite{gella2017image,calixto2017multilingual}. 
Recently \citet{harwath2018vision} also showed the benefit of joint bilingual training in the domain of speech-to-image and image-to-speech retrieval using English and Hindi data.

\section{Multilingual grounded learning}
\begin{figure}
\begin{algorithmic}

\Require $p$: task switching probability.\\

$\mathcal{D}_{c2i}$: datasets $D_1 \dots D_k$ of image-caption pairs \\\hspace{2.5em}$<c, i>$ for all $k$ languages.\\ 

$D_{c2c}$: data set of all possible caption pairs \\\hspace{2.5em}$<c_a, c_b>$ for all $k$ languages. \\

 $\phi (c, \theta_\phi)$: caption encoder\\
$\psi (i, \theta_\phi)$: image encoder
\State
\While {not stopping criterion}

  \State  $T \sim \text{Bern}(p)$
  \If {$T = 1$ }
  	  \State $D_n \sim \mathcal{D}_{c2i}$
      \State $<c, i> \sim D_n$
      \State $\mathbf{a} \gets \phi (c, \theta_\phi)$
      \State $\mathbf{b} \gets \psi (i, \theta_\psi)$
  \Else
      \State $<c_a, c_b> \sim D_{c2c}$
      \State $\mathbf{a} \gets \phi (c_a, \theta_\phi)$
      \State $\mathbf{b} \gets \phi (c_b, \theta_\phi)$
  \EndIf
  \State $[\theta_\phi; \theta_\psi] \gets \text{SGD}(\nabla_{[\theta_\phi; \theta_\psi]} \mathcal{J}(\mathbf{a}, \mathbf{b}))$
\EndWhile
\end{algorithmic}
\caption{Pseudo-code of the training procedure used to train our multilingual multi-task model.}
\label{fig:algo}
\end{figure}

We train a standard model of grounded language learning which  projects images and their textual descriptions into the same space \cite{kiros2014unifying,karpathy2015deep}. 
The training procedure is illustrated by the pseudo-code in Figure~\ref{fig:algo}. 
Images $i$ are encoded by a fixed pre-trained CNN followed by a learned affine transformation $\psi(i, \theta_\psi)$, and captions $c$ are encoded by a randomly initialized RNN $\phi(c, \theta_\phi)$.
The model learns to minimize the distance between pairs $\textrm{<}a,b\textrm{>}$ using a max-of-hinges ranking loss \cite{faghri2017vse++}:
\begin{equation*}
\begin{split}
\mathcal{J}(a, b) = \max_{<\hat{a}, b>}[\text{max}(0, \alpha - s(a,b) + s(\hat{a}, b))] \;+ \\ \max_{<a, \hat{b}>}[\text{max}(0, \alpha - s(a,b) + s(a, \hat{b}))]
\end{split}
\end{equation*}
where $<a, b>$ are the true pairs, and $<a, \hat{b}>$ and $<\hat{a}, b>$ are all possible contrastive pairs in the mini-batch. 
The pairs either consists of image-caption pairs $<i , c>$, where the model solves a caption-image  {\bf c2i} ranking task, or pairs of 
captions in multiple languages belonging to the same image $<c_a, c_b>$, where the model solves a caption-caption {\bf c2c} ranking task \cite{gella2017image}.
Our monolingual models are trained to minimize the caption-image
ranking objective {\bf c2i} on the training set. 
The multilingual models are trained to minimize the ranking loss for the set of all languages $\mathcal{L}$ in the collection: at each iteration the model is either updated for the {\bf c2i} objective or the caption-caption {\bf c2c} objective given either $<c^l, i>$ or a  $<c^k_a, c^m_b>$ pair in languages $l,k,m, \ldots \in \mathcal{L}$. All models are trained by first selecting a task, either {\bf c2i} or {\bf c2c}. In the {\bf c2i} case, a language is sampled at random followed by sampling a random batch; in the {\bf c2c} case, all possible $<c_a, c_b>$ pairs across all languages are treated as a single data set. All of the model parameters are shared across all tasks and languages. 

\paragraph{Implementation.}
We build our model on the PyTorch implementation
of the VSE++ model \cite{faghri2017vse++}.
Images are represented by the 2048D average-pool features
extracted from the ResNet50 architecture \cite{he2016deep} trained on ImageNet \cite{deng2009imagenet}; this is followed by a trained linear layer 
$\mathbf{W}_{I} \in \mathbb{R}^{2048 \times 1024}$. Other implementation details follow \cite{faghri2017vse++}: sentences are represented as the final hidden state of a GRU \cite{chung2014empirical} with 1024 units and 300 dimensional word-embeddings trained from scratch. We use a single word embedding matrix containing  the union of all words in all considered languages.
The similarity function $s$ in the ranking loss is cosine similarity. We $\ell_2$ normalize both the caption and
image representations. The model is trained
with the Adam optimizer \cite{kingma2014adam} 
using default parameters and learning-rate of \mbox{2e-4}. 
We train the model with an early stopping criterion, which is to maximise the 
sum of the image--sentence recall scores R@1, R@5, R@10 on the validation set with patience of 10 evaluations. 
In the monolingual setting the stopping criterion is evaluated at the end of each epoch, whereas in the multilingual setup it is evaluated every 500 iterations. 
The probability of switching between the {\bf c2i} and {\bf c2c} tasks is set to $0.5$. Batches from all data sets are sampled by shuffling the full dataset, going through each batch and re-shuffling when exhausted. 
The sentence-pair dataset used to train the {\bf c2c} ranking model for $\ell$ languages is generated as follows. 
For a given image $i$, a set of languages $1 \cdots$ $\ell$, and a set of captions $C^i_1, \ldots, C^i_\ell$ associated with an image $i$, we generate the set of all possible combinations of size~2 from caption sets $\mathcal{C}^i$ and add the Cartesian product between all resulting pairs $C^i_m \times C^i_n$ in $\mathcal{C}^i$ to the training set.


\section{Experimental setup\footnote{Code to reproduce our results is available at\\ \url{https://github.com/kadarakos/mulisera}.}}\label{sec:data}

\begin{table}
    \centering
    \renewcommand{\arraystretch}{1.3}
	\begin{tabular}{ccccc}
     \toprule
	 & En & De & Fr & Cz \\
     \midrule
     En & 1.0 & 0.04 & 0.06 & 0.02 \\ 
     De & --  & 1.0  & 0.03  & 0.01 \\
     Fr & --  & --   & 1.0   & 0.01 \\
     Cz & --  & --   & --    & 1.0\\
     \bottomrule
	\end{tabular}
   \caption{Vocabulary overlap as measured by the Jaccard coefficient between the different languages on the translation portion of the Multi30K dataset.}\label{tab:data:vocab_overlap}
\end{table}

\paragraph{Datasets.}
We train and evaluate our models on the {\it translation} and {\it comparable} portions of the Multi30K dataset \cite{elliott2016multi30k,elliott2017findings}. The translation portion (a low-resource dataset) contains 29K images, each  described in one English caption with German, French, and Czech translations. The comparable portion (a higher-resource dataset) contains the same 29K images paired with five English and five German descriptions collected independently. Figure \ref{fig:data:example} presents an example of the translation and comparable portions of the data. We used the preprocessed version of the dataset, in which the text is lowercased, punctuation is normalized, and the text is tokenized\footnote{\url{https://github.com/multi30k/dataset}}. To reduce the vocabulary size of the joint models, we replace all words occurring fewer than four times with a special ``UNK'' symbol. Table \ref{tab:data:vocab_overlap} shows the overlap between the vocabularies of the \emph{translation} portion of the Multi30K dataset. The total number of tokens across all four languages is 17,571, and taking the union of the tokens in these four languages results in vocabulary of 16,553 tokens -- a 6\% reduction in vocabulary size. On the \emph{comparable} portion of the dataset, the total vocabulary between English and German contains 18,337 tokens, with a union of 17,667, which is a 4\% reduction in vocabulary size. 

\paragraph{Evaluation.}
We evaluate our models on the 1K images of the 2016 test set of Multi30K either using the 5K captions from the comparable data or the 1K translation pairs. We evaluate on image-to-text (I$\rightarrow$ T) and text-to-image (T$\rightarrow$ I) retrieval tasks. For most experiments we  report Recall at 1 (R@1), 5 (R@5) and 10 (R@10) scores averaged over 10 randomly initialised models. However, in Section~\ref{sec:multi} we only report R@10 due to space limitations and because it has less variance than R@1 or R@5. 





\section{Bilingual Experiments}
\label{sec:bi}

\begin{table*}[t]
\centering
\renewcommand{\arraystretch}{1.3}
\begin{tabular}{llcccccc}
\toprule
& & \multicolumn{3}{c}{I$\rightarrow$T}  & \multicolumn{3}{c}{T$\rightarrow$I}   \\
 						& & R@1 & R@5 & R@10 & R@1 & R@5 & R@10 \\
\cmidrule{3-8}
\parbox[t]{2mm}{\multirow{3}{*}{\rotatebox[origin=c]{90}{\small Symmetric}}} & {\tt VSE}				& 31.6 & 60.4 & 72.7 & 23.3 & 53.6 & \bf{65.8}\\
& {\tt Pivot-Sym}		    & 31.6 & 61.2 & 73.8  & 23.5 & 53.4 & \bf{65.8}\\
& {\tt Parallel-Sym}      & \bf{31.7} & \bf{62.4} & \bf{74.1}  & \bf{24.7} & \bf{53.9} & 65.7 \\
\midrule
\parbox[t]{2mm}{\multirow{3}{*}{\rotatebox[origin=c]{90}{\small Asymmetric}}} & {\tt OE}				& \bf{34.8} & \bf{63.7} & 74.8 & 25.8 & \bf{56.5} & 67.8\\
& {\tt Pivot-Asym}    	& 33.8 & 62.8 & \bf{75.2}  & 26.2 & 56.4 & \bf{68.4}\\
& {\tt Parallel-Asym}     & 31.5 & 61.4 & 74.7  & \bf{27.1} & 56.2 & 66.9\\
\midrule
& Monolingual	  			& 42.4 & 69.9 & 79.8 & 30.5 & 57.8 & 67.9  \\
& Bilingual 	  			& 42.7 & 70.7 & 80.1 & 30.6 & 58.1 & 68.3  \\
& + c2c     	  			& \bf{43.8} & \bf{71.8} & \bf{81.4} & \bf{32.3} &  \bf{59.9} & \bf{70.2} \\
\bottomrule
\end{tabular}
\caption{English Image-to-text (I$\rightarrow$T) and text-to-image (T$\rightarrow$I) retrieval results on the \emph{comparable} part of Multi30K, measured by Recall at 1, 5 at 10. {\tt Typewriter} font shows performance of two sets of symmetric and asymmetric models from \citet{gella2017image}.}
\label{tab:bi:bilingualEng}
\end{table*}

\begin{table*}[h!]
\centering
\renewcommand{\arraystretch}{1.3}
\begin{tabular}{llcccccc}
\toprule
& & \multicolumn{3}{c}{I$\rightarrow$T}  & \multicolumn{3}{c}{T$\rightarrow$I}   \\
 						& & R@1 & R@5 & R@10 & R@1 & R@5 & R@10 \\
\cmidrule{3-8}
\parbox[t]{2mm}{\multirow{3}{*}{\rotatebox[origin=c]{90}{\small Symmetric}}} & {\tt VSE} 			 	& \bf{29.3} & \bf{58.1} & \bf{71.8} & 20.3 & \bf{47.2} & \bf{60.1}\\
& {\tt Pivot-Sym}			& 26.9  & 56.6  & 70.0 & 20.3 & 46.4 & 59.2 \\
& {\tt Parallel-Sym}      & 28.2  & 57.7  & 71.3 & \bf{20.9} & 46.9 & 59.3 \\
\midrule
\parbox[t]{2mm}{\multirow{3}{*}{\rotatebox[origin=c]{90}{\small Asymmetric}}} & {\tt OE} 				& 26.8 & 57.5 & 70.9 & 21.0 & 48.5 & 60.4 \\
& {\tt Pivot-Asym} 		& 28.2 & \bf{61.9} & \bf{73.4} & \bf{22.5} & 49.3 & 61.7 \\
& {\tt Parallel-Asym}     & \bf{30.2} & 60.4 & 72.8 & 21.8 & \bf{50.5} & \bf{62.3} \\
\midrule
& Monolingual				& 34.2 & 63.0 & 74.0 & 23.9 & 49.5 & 60.5\\
& Bilingual 				& 35.2 & 64.3 & 75.3 & 24.6 & 50.8 & 62.0\\
& + c2c     				& \bf{37.9} & \bf{66.1} & \bf{76.8} & \bf{26.6} & \bf{53.0} & \bf{64.0} \\
\bottomrule
\end{tabular}
\caption{German Image-to-text (I$\rightarrow$T) and text-to-image (T$\rightarrow$I) retrieval results on the \emph{comparable} part of Multi30K, measured by Recall at 1, 5 at 10. {\tt Typewriter} font shows performance of two sets of symmetric and asymmetric models from \citet{gella2017image}.}
\label{tab:bi:bilingualGer}
\end{table*}

\subsection{Reproducing \citet{gella2017image}}
\label{sec:gella}

We start by attempting to reproduce the findings of \newcite{gella2017image}. In these experiments we train our multi-task learning model on the {\it comparable} portion of Multi30K. Our models re-implement their setups used for \texttt{VSE} (Monolingual) and bilingual models \texttt{Pivot-Sym} (Bilingual) and \texttt{Parallel-Sym} (Bilingual + c2c). The \texttt{OE}, \texttt{Pivot-Asym} and \texttt{Parallel-Asym} models are trained using the asymmetric similarity measure introduced for the order-embeddings \cite{vendrov2015order}. The main differences between our models and \newcite{gella2017image} is that they use VGG-19 image features, whereas we use ResNet50 features, and we use the max-of-hinges loss instead of the more common sum-of-hinges loss.

Table~\ref{tab:bi:bilingualEng} shows the results on the English comparable 2016 test set. Overall our scores are higher than \newcite{gella2017image}, which is most likely due to the different image features (\newcite{faghri2017vse++} also report a large performance gain when they use the ResNet instead of the VGG image features). Nevertheless, our results show a similar trend to the symmetric cosine similarity models from \newcite{gella2017image}: our best results are achieved with bilingual joint training with the added c2c objective. Their models trained with an asymmetric similarity measure show a different trend: the monolingual model is stronger than the bilingual model, and the c2c loss provides no clear improvement. 

Table~\ref{tab:bi:bilingualGer} presents the German results. Once again, our implementation outperforms \newcite{gella2017image}, and this is likely due to the different visual features and max-of-hinges loss. However, our Bilingual model with the additional c2c objective performs the best for German, whereas \newcite{gella2017image} reports the overall best results for the monolingual baseline \texttt{VSE}. Their models that use the asymmetric similarity function are clearly better than the Monolingual \texttt{OE} model. In general, the results from \newcite{gella2017image} indicate the benefits of bilingual joint training, however, they do not find a clear pattern between the model configurations across languages. In our implementation, we only focused on the symmetric cosine similarity function and found a systematic pattern across both languages: bilingual training improves results on all performance metrics for both languages, and the additional c2c objective always provides further improvements.

\subsection{Translations vs.\ independent captions}
\label{sec:bitrans}

\begin{table}[t]
\centering
\renewcommand{\arraystretch}{1.3}
\begin{tabular}{lcccc}
\toprule
& \multicolumn{2}{c}{English} &  \multicolumn{2}{c}{German} \\
 & I$\rightarrow$T & T$\rightarrow$I & I$\rightarrow$T & T$\rightarrow$I \\
\midrule
Monolingual			& 56.3 & 40.1 & 39.5 &  20.9\\
\midrule
Bi-translation 			& 67.4 &  55.1 & 58.3 & 44.6 \\
+ c2c     & 58.2 & 47.7 & 51.0 & 39.6\\
\midrule
Bi-comparable        & {\bf 67.9} & 55.7 & {\bf 62.0} & 48.1\\
+ c2c  & 67.6 & {\bf 56.0} & 61.9 & {\bf 49.1} \\
\bottomrule
\end{tabular}
\label{tab:bilingual}
\caption{R@10 retrieval results on the \emph{comparable} part of Multi30K. Bi-translation is trained on 29K \emph{translation pair} data; bi-comparable is trained by downsampling the \emph{comparable} data to 29K. 
}
\label{tab:bitrans}
\end{table}

We now study whether the model can be trained on either translation pairs or independently collected bilingual captions. \newcite{gella2017image} only conducted experiments on independently collected captions. However, it is known that humans have equally strong preference for translated or independently collected captions of images \cite{frank_elliott_specia_2018}, which has implications for the difficulty and cost of collecting training data. Our baseline is a Monolingual model trained on 29K single-captioned images in the {\it translation} portion of Multi30K. The Bi-translation model is trained on both German and English, with shared parameters. Table~\ref{tab:bitrans} shows that there is a substantial improvement in performance for both languages in the bilingual setting. However, the additional c2c loss degrades performance here. This could be because we only have one caption per image in each language and it is easier to find a relationship between these views of the translation pairs. 

In the Bi-comparable setting, we randomly select an English and a German sentence for each image in the {\it comparable} portion of Multi30K. We only find a minor difference in performance between the Bi-translation and Bi-comparable models for English, but the German results are improved. Crucially, it is still better than training on monolingual data. In the Bi-comparable setting, the c2c loss does not have a detrimental effect on model performance, unlike in the Bi-translation experiment. Overall we find that the \emph{comparable} data leads to larger improvements in retrieval performance.

\subsection{Overlapping vs.\ non-overlapping images}
\label{sec:bioverlap}

\begin{table}
\renewcommand{\arraystretch}{1.3}
\begin{tabular}{lcccc}
\toprule
& \multicolumn{2}{c}{English} &  \multicolumn{2}{c}{German} \\
 & I$\rightarrow$T & T$\rightarrow$I & I$\rightarrow$T & T$\rightarrow$I \\
\midrule
Full Monolingual			& 79.8 & 67.9 & 74.0 & 60.5\\
Half Monolingual			& 73.7 & 61.6 & 66.4 & 53.9 \\
\midrule
Bi-overlap				& 73.6 & 62.2 & 67.6 &  54.9 \\
+ c2c 		& \bf{76.0} & \bf{65.9} & \bf{71.2}  & \bf{59.1}\\
\midrule
Bi-disjoint 	& 73.1 & 62.1 & 67.9 &  54.9\\
\bottomrule
\end{tabular}
\caption{R@10 retrieval results on the \emph{comparable} part of Multi30K. Full model trained on the 29K images of the \emph{comparable} part, Half model on 14.5K images using random downsampling. For Bi-overlap, both English and German captions are used for 14.5K images. For Bi-disjoint, 14.5K images are used for English and the remaining 14.5K images for German.}
\label{tab:half}
\end{table}

In a bilingual setting, we can improve an image-sentence ranking model by collecting more data in a second language. This can be achieved in two ways:  by collecting captions in a new language for the same overlapping set of images, or by  using a disjoint set of images and captions in a new language. We compare these two settings here. 

In the Bi-overlap condition, we collect captions for the existing images in a new language, i.e.\ we use all of the English and German captions paired with a random selection of 50\% of the images in {\it comparable} Multi30K. This results in a training dataset of 14.5K images with 145K bilingual captions. In the Bi-disjoint condition, we collect captions for new images in a new language, i.e.\  we use all of the English captions from a random selection of 50\% of the images, and all of the German captions for the remaining 50\% of the images. This results in a training dataset on 29K images with a total of 145K bilingual captions. 

Table~\ref{tab:half} shows the results of this experiment. The upper-bound is to train a Monolingual model on the full \textit{comparable} corpus. For the lower bound, we train Half Monolingual models by randomly sampling half of the 29K images and their associated captions, giving 72.5K captions over 14.5K images. Unsurprisingly, the Half Monolingual models perform worse than the Full Monolingual models. In the Bi-overlap experiment, the German model is improved by collecting captions for the existing images in English. There is no difference in the performance of the English model, echoing the results from Section~\ref{sec:gella}. The Bi-overlap model also benefits from the added c2c objective. Finally, the Bi-disjoint model performs as well as the Bi-overlap model without the c2c objective. (It was not possible to train the Bi-disjoint model with the additional c2c objective because there are no caption pairs for the same image.)

Overall, these results suggest that it is best to collect additional captions in the original language, but when adding a second language, it is better to collect extra captions for existing images and exploit the additional c2c ranking objective.

\section{Multilingual experiments}
\label{sec:multi}

We now turn out attention to multilingual learning using the English, German, French and Czech annotations in the {\it translation} portion of Multi30K. We only report the text-to-image (T$\rightarrow$I) R@10 results due to space limitations. 

We did not repeat the overlapping vs.\ non-overlapping experiments from Section~\ref{sec:bioverlap} in a multilingual setting because this would introduce too much data sparsity. In order to conduct this experiment, we would have to downsample the already low-resource French and Czech captions by 50\%, or even further for multi-way experiments.

\subsection{Translation vs.\ independent captions}
\label{sec:multitrans}

Table~\ref{tab:multilingual} shows the results of repeating the translations vs.\ comparable captions experiment from Section~\ref{sec:bitrans} with data in four languages. The Multi-translation models are trained on 29K images paired with a single caption in each language. These models perform better than their Monolingual counterparts, and the German, French, and Czech models are further improved  with the c2c objective. The Multi-comparable models are trained by randomly sampling one English and one German caption from the {\it comparable} dataset, alongside the French and Czech translation pairs. These models perform as well as the Multi-translation models, and the c2c objective brings further improvements for all languages in this setting.

These results clearly demonstrate the advantage of jointly training on more than two languages. Text-to-image retrieval performance increases by more than 11 R@10 points for each of the four languages in our experiment.

\subsection{High-to-low resource transfer}
\label{sec:multitransfer}

\begin{table}[t]
\centering
\renewcommand{\arraystretch}{1.3}
\begin{tabular}{lcccc}
\toprule
& En & De & Fr & Cz \\
\cmidrule{2-5}
Monolingual & 50.4 & 39.5 & 47.0 & 42.0 \\
\midrule
Multi-translation & 58.7 & 51.2 &  57.0 & 51.0\\
+ c2c & 56.3  & 52.2 & 55.0 & 51.6\\
\midrule
Multi-comparable & 59.2 & 49.6 & 57.2 & 50.8 \\
 + c2c & \bf{61.8}  & \bf{52.7} & \bf{59.2} & \bf{55.2} \\
\bottomrule
\end{tabular}
\caption{The Monolingual and joint Multi-translation models trained on \emph{translation pairs}, and the Multi-comparable trained on the downsampled \emph{comparable} set with one caption per image.}
\label{tab:multilingual}
\end{table}

We now examine whether the lower-resource French and Czech models benefit from training with the full complement of the higher-resource English and German comparable data. Therefore we train a joint model on the \textit{translation} as well as \textit{comparable} portions of Multi30K, and examine the performance on French and Czech.

Table~\ref{tab:multilingual-highlow} shows the results of this experiment. We find that the French and Czech models improve by 8.8 and 5.5 R@10 points respectively when they are only trained on the multilingual translation pairs (compared to the monolingual version), and by another 2.2 and 2.8 points if trained on the extra 155K English and German \textit{comparable} descriptions. We also find that the additional c2c objective improves the Czech model by a further 4.8 R@10 points (this improvement is likely caused by training the model on 46 possible caption pairs). Our results show the impact of jointly training with the larger English and German resources, which demonstrates the benefits of high-to-low resource transfer.

\subsection{Bilingual vs. multilingual}
\label{sec:bivsmult}

\begin{table}
\centering
\renewcommand{\arraystretch}{1.3}
\begin{tabular}{lcc}
\toprule
& French & Czech \\
\cmidrule{2-3}
Monolingual    & 47.0 & 42.0\\
Multilingual   & 56.3 & 51.3 \\
+ Comparable   & 58.9 & 52.4 \\
+ c2c          & \bf{61.6} & \bf{57.2} \\
\bottomrule
\end{tabular}
\caption{Multilingual is trained on all \emph{translation pairs}, 
+ Comparable adds the \emph{comparable} data set.}
\label{tab:multilingual-highlow}
\end{table}

\begin{figure*}
\begin{center}
\begin{tikzpicture}[yscale=1.1, xscale=1.05]

\begin{groupplot}[
   group style={
       group size=2 by 1,
       x descriptions at=edge bottom,
       y descriptions at=edge left,
       vertical sep=0pt,
       horizontal sep=15pt},
  ybar,
  x=2.3cm,
  every node near coord/.append style={font=\normalsize},
  nodes near coords,
  nodes near coords align=vertical,
  enlarge x limits=0.2,
  point meta=y * 1, 
  ylabel={\textbf{R@10}},
  tick align=outside,
  ytick={0,20,...,100}, 
  xtick={1,...,3},
  xticklabels={
    Monolingual,
    Bilingual,
    Multilingual},
  x tick label style={align=left},
  ymin=0,
  ymax=100
]

\nextgroupplot[bar width=18pt]
\addplot[draw=red,fill=red!50]
coordinates {(1,56.3) (2,67.6) (3,71.9)};
\addplot[draw=blue, fill=blue!50] 
coordinates {(1,40.1) (2,56.0) (3,61.0)};
\node at (rel axis cs:0.5,0.9) {English};

\nextgroupplot[bar width=18pt,legend style={at={(-0.05,1)},anchor=north}]
\addplot[draw=red,fill=red!50]
coordinates {(1,39.5) (2,61.9) (3,65.1)};
\addlegendentry{Image $\rightarrow$ Text}
\addplot[draw=blue, fill=blue!50] 
coordinates {(1,20.9) (2,49.1) (3,52.6)};
\addlegendentry{Text $\rightarrow$ Image}

\node at (rel axis cs:0.5,0.9) {German};
\end{groupplot}
\end{tikzpicture}
\caption{Comparing models from the Monolingual,  Bilingual and Multilingual settings. The Monolingual and Bilingual models are trained on the downsampled English and German \emph{comparable} sets with additional c2c objective. The Multilingual model uses the French and Czech \emph{translation pairs} as additional data. The results are reported on the full 2016 test set of the \emph{comparable} portion of Multi30K.}
\label{fig:chart}
\end{center}
\end{figure*}
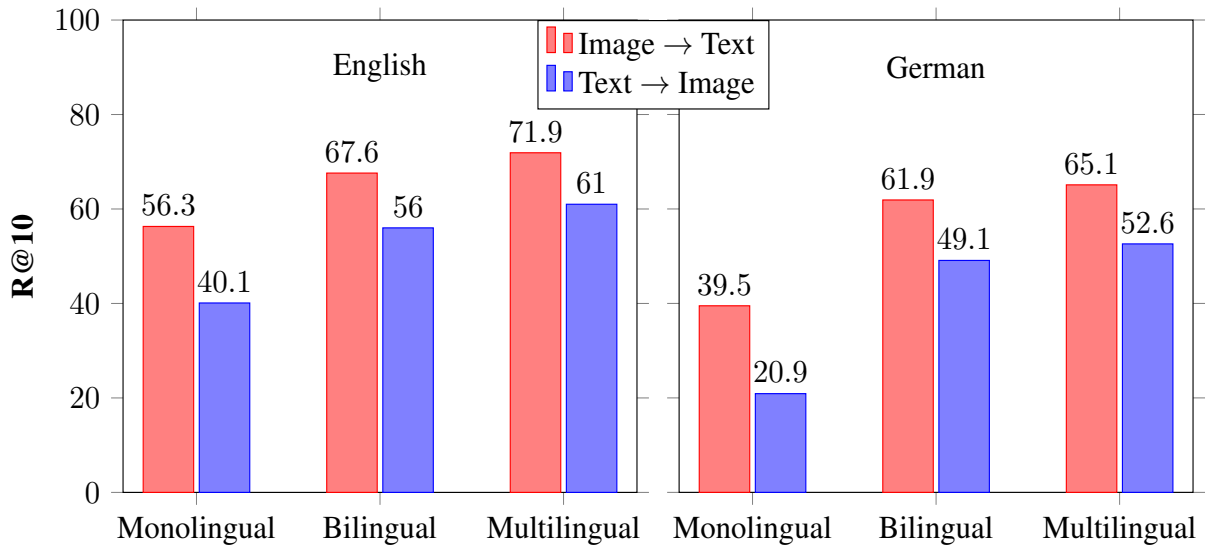

Finally, we investigate how useful it is to train on four languages instead of two. Figure~\ref{fig:chart} presents the image-to-text and text-to-image retrieval results of training Monolingual, Bilingual, or Multilingual models. The Monolingual and Bilingual models are trained on a random single-caption-image subsample of the {\it comparable} dataset with the additional c2c objective, as this configuration provided the overall best results in Sections~\ref{sec:bitrans} and \ref{sec:multitrans}. The Multilingual models are trained with the additional French and Czech {\it translation} data. As can be seen in Figure~\ref{fig:chart}, the performance on both tasks and for both languages improves as we move from using data from one to two to four languages.

\section{Conclusions}
We learn multilingual multimodal sentence embeddings and show that multilingual joint training improves over bilingual joint training. We also demonstrate that low-resource languages can benefit from the additional data found in high-resource languages. Our experiments suggest that either translation pairs or independently-collected captions improve the performance of a multilingual model, and that the latter data setting provides further improvements through a caption--caption ranking objective. We also show that when collecting data in an additional language, it is better to collect captions for the existing images because we can exploit the caption--caption objective. Our results lead to several directions for future work. We would like to pin down the mechanism via which multilingual training contributes to improved performance for image-sentence ranking. Additionally, we only consider four languages and show the gain of multilingual over bilingual training only for the English-German language pair. In future work we will incorporate more languages from data sets such as the Chinese Flickr8K \cite{li2016adding} or Japanese COCO \cite{P16-1168}.

\section*{Acknowledgements}

Desmond Elliott was supported by an Amazon Research Award.

\bibliographystyle{apalike}
\bibliography{biblio}

\end{document}